\documentclass[letterpaper]{article} % DO NOT CHANGE THIS
\usepackage{aaai2026}  % DO NOT CHANGE THIS
\usepackage{times}  % DO NOT CHANGE THIS
\usepackage{helvet}  % DO NOT CHANGE THIS
\usepackage{courier}  % DO NOT CHANGE THIS
\usepackage[hyphens]{url}  % DO NOT CHANGE THIS
\usepackage{graphicx} % DO NOT CHANGE THIS
\usepackage{natbib}  % DO NOT CHANGE THIS AND DO NOT ADD ANY OPTIONS TO IT
\usepackage{caption} % DO NOT CHANGE THIS AND DO NOT ADD ANY OPTIONS TO IT
\usepackage{algorithm}
\usepackage{algorithmic}
\usepackage{amsfonts}

\usepackage{pgfplots}
\usepackage{subcaption}
\usepackage{booktabs}

\usepackage{newfloat}
\usepackage{listings}
\usepackage{multirow}
\DeclareCaptionStyle{ruled}{labelfont=normalfont,labelsep=colon,strut=off} % DO NOT CHANGE THIS
\floatstyle{ruled}
\newfloat{listing}{tb}{lst}{}
\floatname{listing}{Listing}
\title{Few-Shot Precise Event Spotting via Unified Multi-Entity Graph and Distillation}
\author{
    Zhaoyu Liu\textsuperscript{\rm 1}, 
    Kan Jiang\textsuperscript{\rm 1}, 
    Murong Ma\textsuperscript{\rm 1}, 
    Zhe Hou\textsuperscript{\rm 2},
    Yun Lin\textsuperscript{\rm 3},
    Jin Song Dong\textsuperscript{\rm 1}
}
\affiliations{
    \textsuperscript{\rm 1}National University of Singapore \quad
    \textsuperscript{\rm 2}Griffith University \quad
    \textsuperscript{\rm 3}Shanghai Jiao Tong University\\
    \texttt{\{liuzy, jiangkan\}@nus.edu.sg, murongma@u.nus.edu}\\
    \texttt{z.hou@griffith.edu.au, lin\_yun@sjtu.edu.cn, dcsdjs@nus.edu.sg}
}

\usepackage{bibentry}
\newcommand{\model}[0]{\text{UMEG-Net}}

\usepackage{bibentry}
\usepackage{amsmath}

\newcommand{\rev}[1]{{\color{black}#1}}

\pgfplotsset{compat=1.18}
\begin{document}

\maketitle

\begin{abstract}
Precise event spotting (PES) aims to recognize fine-grained events at exact moments and has become a key component of sports analytics. This task is particularly challenging due to 
%factors such as 
rapid succession, motion blur, and subtle visual differences. Consequently, most existing methods rely on domain-specific, end-to-end training with large labeled datasets and often struggle in few-shot conditions due to their dependence on pixel- or pose-based inputs alone. 
However, obtaining large labeled datasets is practically hard.
We propose a Unified Multi-Entity Graph Network (\model{}) for few-shot PES.
% , further enhanced by a multimodal distillation module to improve robustness in challenging scenarios. 
\model{} integrates human skeletons and sport-specific object keypoints into a unified graph and features an efficient spatio-temporal extraction module based on advanced GCN and multi-scale temporal shift. To further enhance performance, we employ multimodal distillation to transfer knowledge from keypoint-based graphs to visual representations. Our approach achieves 
% \rev{[put numbers]} 
robust performance with limited labeled data and significantly outperforms baseline models in few-shot settings, providing a scalable and effective solution for few-shot PES. Code is publicly available at \url{https://github.com/LZYAndy/UMEG-Net}.
\end{abstract}


\section{Introduction}

Precise Event Spotting (PES) is a trending problem that aims to identify events and their class from long, untrimmed videos, particularly in sports~\cite{hong2022spotting,xarles2024t,xu2025actionspottingpreciseevent}.
% that aims to identify the exact moment a specific event occurs within long, untrimmed videos~\cite{xu2025actionspottingpreciseevent}, particular in sports. 
The main objective is to accurately detect sequences of fine-grained, rapidly occurring sports events, such as a hitting event in racket sports, within a tight tolerance window (1-2 frames). This capability is essential for sports analytics applications 
such as match forecasting~\cite{wang2022shuttlenet,liu2025analyzingnba}, strategic and tactical analysis~\cite{dong2023sports,liu2023insight,liu2024exploring,liu2024strategy,liu2024pcsp,liu2025analyzing}, and player performance evaluation~\cite{decroos2019actions,pappalardo2019playerank}.
% such as action forecasting~\cite{wang2022shuttlenet}, strategic and tactical analysis~\cite{liu2023insight,dong2023sports}, and player performance evaluation~\cite{decroos2019actions}.
Most existing PES methods are trained end-to-end using raw RGB images as input~\cite{hong2022spotting,xarles2024t,liu2025f}.
One main challenge with these approaches is that they rely on large-scale datasets with dense, frame-level annotations~\cite{shao2020finegym,xu2022finediving,wang2023shuttleset,liu2025f}, which leads to substantial labeling costs and significantly limits their scalability to new domains~\cite{xu2025actionspottingpreciseevent}.
% \rev{These approaches typically require large-scale datasets with dense, frame-level annotations~\cite{shao2020finegym,xu2022finediving,wang2023shuttleset,bian2024p2anet,liu2025f}, resulting in high labeling costs and limited scalability to new domains \cite{}. }
% Furthermore, reliance on pixel-level features often leads to overfitting to static visual patterns rather than learning generalizable motion representations~\cite{choi2019can,weinzaepfel2021mimetics}, particularly in low-data scenarios.

\begin{figure}
\centering
    \includegraphics[width=\linewidth]{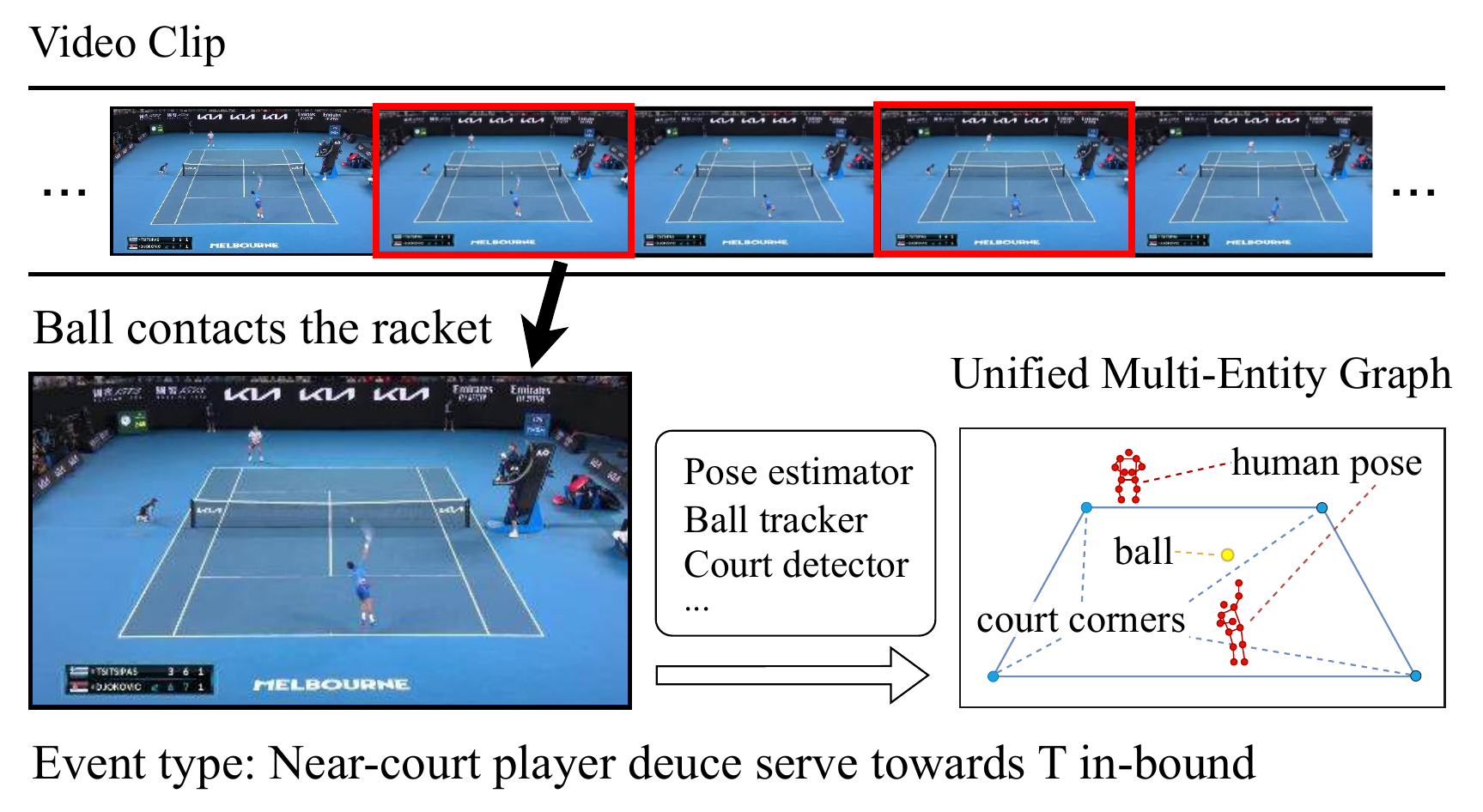}
\caption{Precise event spotting in sports videos, with event timestamps highlighted in red. Each scene can be represented by a unified graph including human poses and sport-related entity keypoints (e.g., ball, court corners).}
\label{fig:vid2seq}
\end{figure}

These limitations underscore the significance of \emph{few-shot learning for PES}, an area that remains under-explored in existing research. Traditional few-shot methods, such as contrastive~\cite{chen2020simple} and meta-learning~\cite{wanyan2025comprehensivereviewfewshotaction} approaches, typically excel at coarse-grained, action-level recognition but do not directly address the frame-level accuracy required by PES. Recent advancements in human pose estimation~\cite{sun2019deep} and sports object tracking~\cite{huang2019tracknet,jiang2020deep,jiang2023court,chen2023tracknetv3,jiang2024tracking} 
% (e.g., balls, players, courts) 
have facilitated transforming raw RGB videos into compact, keypoint-based representations suitable for few-shot scenarios. Although numerous studies leverage human pose data for action recognition~\cite{duan2022pyskl,zhou2024blockgcn,liu2025revealing}, these methods usually neglect critical contextual cues from objects (e.g., the ball) and the environment (e.g., the court), resulting in substantial information loss and compromised event detection performance. Some works have attempted to include additional context~\cite{ibh2023tempose,li2024skeleton}, yet they predominantly focus on coarse-grained recognition tasks and thus lack the fine-grained temporal precision essential for PES, making direct comparisons difficult. Additionally, keypoint-based approaches depend on accurate detections, which are often unreliable in fast-moving sports due to motion blur and occlusion. Thus, relying solely on keypoints is suboptimal for robust PES.

To address these limitations, we propose a graph-based method for precise event spotting that flexibly incorporates human skeletons and object-level keypoints (e.g, ball positions and court corners). We first use pose estimation and object detection to extract keypoints from all relevant entities.
% \zhe{Somebody will ask: So how can the method be applied to a different sport? What's needed? Definition of different contextual environment?}. 
These keypoints are integrated into a unified multi-entity graph (see Figure~\ref{fig:vid2seq}), capturing interactions among multiple players, objects, and contextual landmarks, thus overcoming the limitations of standard skeleton-only representations. 
% We construct a unified multi-entity graph (see Figure~\ref{fig:vid2seq}) that integrates multiple human skeletons with object and environment landmarks, overcoming the constraints of standard skeleton representations by supporting multi-person interactions and embedding rich contextual information. 
To efficiently encode spatial-temporal relationships, we introduce the Unified Multi-Entity Graph Network (\model{}), a novel graph-based module explicitly designed for few-shot PES. \model{} employs an advanced Graph Convolutional Network (GCN) backbone for spatial modeling, replacing traditional temporal convolutions with a parameter-free multi-scale temporal shift mechanism. This temporal shift operation effectively captures temporal dynamics without introducing extra parameters, ensuring computational efficiency and robust few-shot performance.
% To efficiently and effectively model spatial-temporal relationships, we introduce Unified Multi-Entity Graph Network (\model{}), a novel graph-based module for few-shot PES. Our design balances expressiveness and computational efficiency, enabling the network to capture insightful spatial and temporal dependencies while remaining lightweight and sample-efficient. 
% % \zhe{Is computational efficiency or real-time processing a highlight? If so, discuss computational issues as challenges in the 2nd paragraph, and highlight that we can do it live/real-time here.} 
% Specifically, we adopt an advanced graph convolutional network (GCN) backbone for spatial modeling and replace conventional multi-stage temporal convolution with a multi-scale temporal shift mechanism. This temporal shift operation introduces no additional parameters, yet efficiently captures temporal correlations across frames. 
Furthermore, to mitigate inaccuracies from keypoint detection, we utilize multimodal knowledge distillation, transferring learned features from the graph-based teacher model to an RGB-based student network (denoted as \model{}$_\text{ distill}$). By leveraging abundant unlabeled videos from the target domain, the student network learns complementary visual representations, enhancing robustness and generalization in few-shot PES scenarios.
% and enabling rapid adaptation to new sports without extensive manual annotations.
% To further address keypoint detection inaccuracies, we employ a multimodal distillation framework to transfer the knowledge from the graph-based teacher network to a visual-based student network. This enables the student to learn complementary features from RGB data, enhancing robustness. By leveraging abundant unlabeled target-domain videos, the student network can adapt to new sports without additional manual annotations, thus improving generalization in few-shot PES scenarios.

\model{} improves few-shot PES across diverse sports domains. We validate its effectiveness on five sports video datasets with fine-grained event types and precise timestamps: F$^3$Set~\cite{liu2025f}, ShuttleSet~\cite{wang2023shuttleset}, FineGym~\cite{shao2020finegym}, Figure Skating~\cite{hong2021video}, and SoccerNet Ball Action Spotting~\cite{cioppa2024soccernet2024challengesresults_short}. Under few-shot conditions, \model{} consistently outperforms baseline methods, improving F1 scores by 1.3\% to 5.5\% and edit scores by 1.3\% to 16.4\%.. Additionally, incorporating multimodal distillation, our visual-based student model $\model{}_{\text{distill}}$ achieves an additional average gain of 5.8\% in F1 score and 6.7\% in edit score,
% further improves performance, with an average improvement of 5.8\% in F1 score and 6.7\% in edit score, 
highlighting the robustness gained from complementary RGB features.
% , which improves edit scores by 13.7\% to 37.8\% over RGB, 14\% to 28\% over optical flow, and 4.3\% to 17.6\% over skeleton-based approaches. 
% Even when labels are plentiful, \model{} remains competitive, achieving state-of-the-art performance across all datasets. 
The key contributions are as follows:
\begin{itemize}
    \item We introduce and investigate the \emph{few-shot} precise event spotting (PES) task, targeting frame-level event recognition with limited labeled data.
    \item We designed a unified multi-entity graph that incorporates human skeletons, objects (e.g., ball), and contextual landmarks (e.g., court corners) to represent sports events.
    % , enabling flexible modeling of multi-person interactions and sport-dependent structures across different domains.
    \item We propose \model{}, a new graph-based framework for few PES. It combines spatial graph convolution with a parameter-free multi-scale temporal shift mechanism, and enhances robustness through multimodal distillation.
    % further enhanced by multimodal distillation for improved generalization and robustness under few-shot conditions.
    % \item We propose \model{}, a unified multi-entity graph network for few-shot PES that flexibly integrates human skeletons and object-level keypoints. It employs efficient spatio-temporal modeling via GCN and multi-scale temporal shift, and enhances robustness through multimodal knowledge distillation from a graph-based teacher to an RGB-based student network.
    
    % We propose a unified multi-entity graph framework that flexibly integrates human skeletons and sport-specific object keypoints into a unified graph for efficient and expressive spatio-temporal feature extraction.
    % \item We develop a multimodal distillation framework that transfers knowledge from the graph-based teacher to a visual-based student network, further enhancing robustness in challenging scenarios.
    \item We conduct extensive experiments and ablation studies across five sports datasets, demonstrating that \model{} achieves state-of-the-art performance in both few-shot and fully supervised settings.
\end{itemize}

\section{\rev{Related Work}}

\subsection{Precise Event Spotting} 
Precise event spotting in sports video analysis, initially introduced by~\cite{einfalt2019frame} for athletics (long and triple jumps), involves identifying exact timestamps of specific actions within strict frame-level tolerances. Subsequently, the SoccerNet dataset~\cite{giancola2018soccernet} expanded this research by providing extensive soccer video annotations for temporal action localization. Recent advances~\cite{hong2022spotting,xarles2024t} have broadened the applicability of PES across various sports. E2E-Spot~\cite{hong2022spotting}, employing a CNN backbone with Gated Shift Modules~\cite{sudhakaran2020gate} and GRUs~\cite{dey2017gate} for temporal modeling, performed effectively in tennis and figure skating scenarios but struggled with the complex temporal dynamics of SoccerNet V2~\cite{cioppa2024soccernet2024challengesresults_short}. Addressing these limitations, T-DEED~\cite{xarles2024t} improved temporal precision for fast-paced sports, achieving state-of-the-art accuracy on Figure Skating and Fine Diving benchmarks. More recently, Liu et. al. introduced a large-scale dataset spanning multiple sports and their proposed F$^3$ED further improved PES performance through efficient visual encoding and contextual sequence refinement~\cite{liu2025f}.
% Another line of research by [35] leveraged ensembles of Kinetics-400 pre-trained networks, generating rich feature sets adopted by subsequent works, including Soares et al. [24] who integrated these features into a U-Net and transformer-based pipeline, and ASTRA [29], which additionally incorporated audio modalities. COMEDIAN [6] diverged from the use of pre-extracted features, employing a multi-stage training approach directly on raw video frames. UGLF [27] introduced vision-language guided separation of global context and localized features to enhance interpretability. 
Despite these advances, PES methods often suffer from limited supervision due to the high cost of fine-grained annotation. Our method addresses these limitations, demonstrating superior event spotting accuracy, particularly in few-shot scenarios.

\subsection{Skeleton-Based Action Understanding} 
Skeleton-based human pose representations have been extensively utilized in sports analytics due to their efficiency, reduced complexity, and robustness in fine-grained action recognition, especially under few-shot conditions. Recent methods leveraging skeleton data have demonstrated effectiveness on single-athlete sports datasets, such as gymnastics (FineGym), achieving strong performance through dedicated skeleton-based frameworks~\cite{duan2022pyskl,liu2025revealing}.
% PoseConv3D (Duan et al., CVPR 2022), ProtoGCN (Wei et al., arXiv 2022), and SkeletonMAE (Yan et al., ICCV 2023). 
Hong et al.~\cite{hong2021video} further highlighted the potential of pose features by distilling pose knowledge into RGB-based networks, illustrating how pose alignment can enhance recognition accuracy, yet they still focus primarily on single-athlete scenarios and remain reliant on accurate pose detection. In team sports such as volleyball and basketball, skeleton-based methods have explored group activity recognition through modeling multi-person interactions~\cite{perez2022skeleton,zhou2022composer}. Recent studies have additionally integrated non-human entities, such as the shuttlecock trajectory and court locations in badminton~\cite{liu2022monotrack,ibh2023tempose}, to enrich contextual information. Liu et al.~\cite{li2024skeleton} improved upon these efforts by constructing panoramic graphs that integrate multiple players and objects but overlooked critical spatial relationships involving court locations. Furthermore, these approaches typically address action recognition tasks without considering precise temporal localization in PES. Additionally, our graph-based module efficiently supports precise temporal spotting and few-shot learning. By further incorporating distillation from raw RGB inputs, our framework remains robust against inaccuracies in pose and object detection, significantly outperforming previous approaches under strict temporal localization constraints and limited data settings.

\begin{figure*}
\centering
    \includegraphics[width=\linewidth]{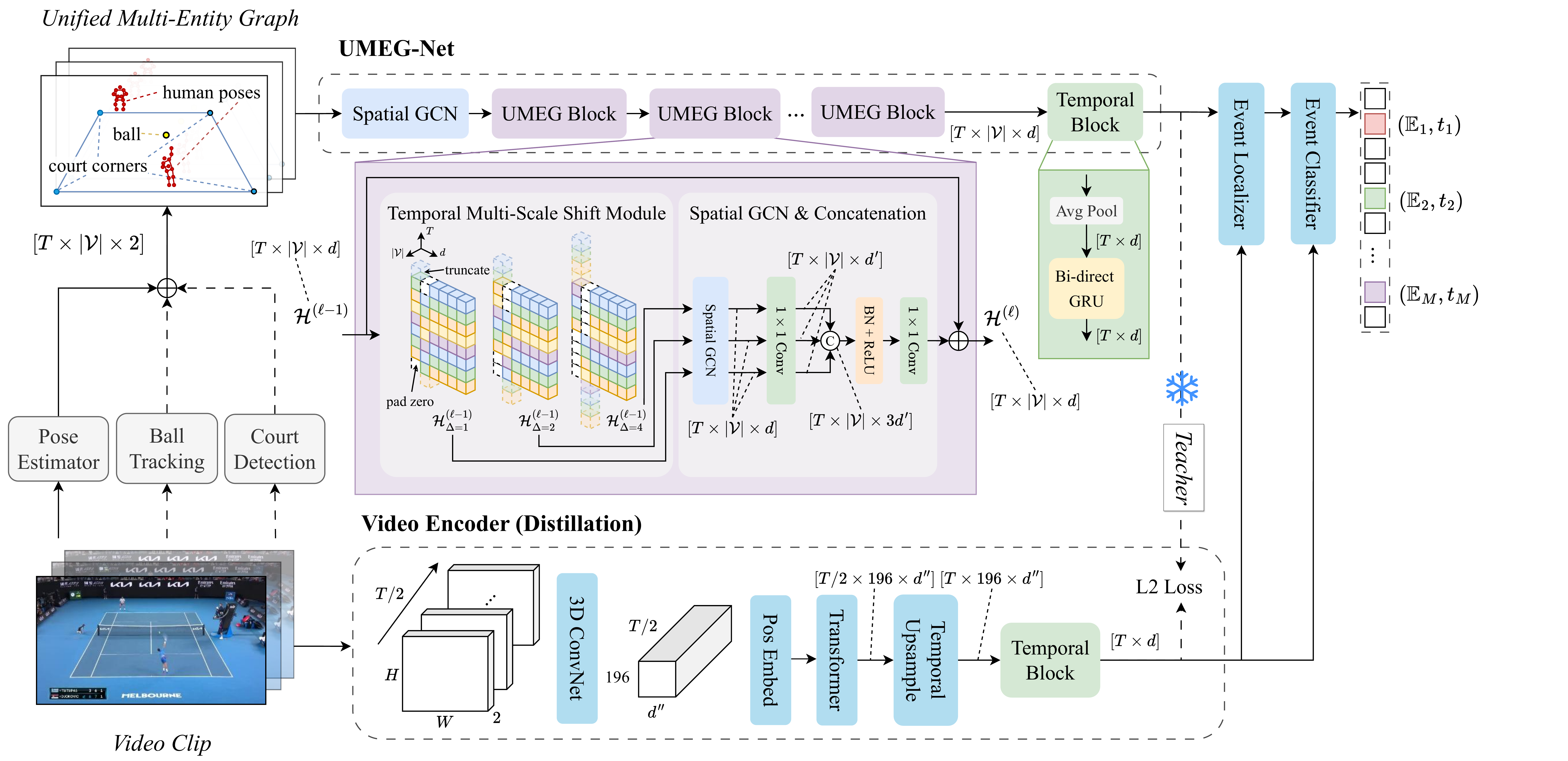}
\caption{\rev{The framework of our proposed method, including UMEG-Net and multimodal distillation. Each frame is converted to a unified multi-entity graph and processed by stacked UMEG Blocks to produce features for precise event spotting. A transformer-based RGB student is trained via knowledge distillation from the frozen graph-based teacher.}}
\label{fig:arch}
\end{figure*}

\section{Proposed Method}

This section presents our method in detail, starting with the problem statement and then presenting the modules.
%as follows. 

\subsection{Problem Statement} 
We define the few-shot PES task as follows:
Given the input video clip $X \in \mathbb{R}^{T \times H \times W \times 3}$ consisting of $T$ frames of RGB image size $H \times W$ with channel size of 3, our goal is to detect a sequence of $M$ event-timestamp pairs $((E_1, t_1), \ldots, (E_M, t_M))$. Here, $E_i$ denotes the event type with $C$ possible classes, and $t_i$ is the corresponding timestamp for $i \in \{1, \ldots, M\}$.  
We consider a target dataset $D$ composed of $|D|$ video \emph{clips}, each containing a certain number of events. Let $D_{\text{label}} \subset D$ denote the subset of $D$ that contains labels, and let $D_{\text{unlabel}} \subset D$ be the unlabeled subset, such that $D = D_{\text{label}} \cup D_{\text{unlabel}}$. 
The number of clips in each subset is $|D_{\text{(un)label}}|$.
% The number of samples (clips) in each subset is denoted as $|D_{\text{(un)label}}|$. 
%\emph{
We specifically address the \textbf{few-shot} scenario, where the number of labeled samples $k = |D_{\text{label}}|$ is small, referred to as the ``k-clip'' setting.
%}
% \emph{Our few-shot scenario refers to cases where $k=|D_{\text{label}}|$ is small, denoted as ``k-clip''.}

\subsection{Our Proposed Framework}

To tackle the problem described above, we propose \model{}, a unified multi-entity graph network explicitly designed for few-shot PES. The overall architecture is depicted in Figure~\ref{fig:arch}. 
% \model{} integrates keypoints representing human skeletons and sport-related entities (e.g., ball, court) into a unified panoramic graph. 
A unified panoramic graph is first constructed by integrating keypoints representing human skeletons and sport-related entities (e.g., ball, court). 
This graph is then processed by a series of UMEG Blocks, each combining advanced spatial graph convolutional networks (GCNs) with a parameter-free multi-scale temporal shift mechanism for efficient spatio-temporal feature extraction.  Finally, the feature maps are forwarded to the event localizer and classifier for precise event spotting.
% It then processes this graph through a set of efficient spatio-temporal feature extraction modules (UMEG Block), each of which contains an advanced Spatial Graph Convolutional Networks (GCNs) and a parameter-free multi-scale temporal shift mechanism. 
To further enhance model robustness and generalization, we introduce a multimodal distillation approach, transferring knowledge from the graph-based model to an RGB-based student network through weak-supervision even on large amounts of unlabeled videos.

\subsubsection{Unified Multi-Entity Graph Construction} 
Given a sports video, we first convert each frame into a structured graph of interacting entities. Specifically, for each frame $t$, we
define a graph $\mathcal{G}_t=(\mathcal{V}_t,\mathcal{E}_t)$, where $\mathcal{V}_t$ is the set of nodes that includes all detected human joints and keypoints of sports-related entities and $\mathcal{E}_t$ is the set of edges. $\mathcal{V}$ including all detected human joints and keypoints of sports-related entities, denoted $\mathcal{V}_t=\{V_p^t,V_b^t,V_c^t\}$, where $V_p^t=\{P_i^t|{i=1,...,N}\}$ is the set of $N$ persons each $P_i^t=(j_{i,1}^t,\dots,j_{i,K}^t)$ represented by joints $K$, 
$V_b^t$ are ball keypoints and $V_c^t$ are court keypoints.
The number of nodes is denoted as $|\mathcal{V}_t| = N*K + |V_b^t| + |V_c^t|$.
The edge set $\mathcal{E}_t$ captures both intra-entity structures and cross-entity interactions. \rev{It contains four components:}
\[
\mathcal{E}_t \;=\; \mathcal{E}^{\text{intra}}_t \;\cup\; \mathcal{E}^{\text{p--b}}_t \;\cup\; \mathcal{E}^{\text{p--c}}_t \;\cup\; \mathcal{E}^{\text{c--c}}_t,
\]
where $\mathcal{E}^{\text{intra}}_t$ encodes skeletal connections within each player (intra-person edges follow standard human-joint topology); $\mathcal{E}^{\text{c--c}}_t$ connects 4 court corners as a rectangle to model the field boundary; $\mathcal{E}^{\text{p--b}}_t$ connects human joints
% (e.g., hands in racket sports or feet in soccer) 
to the ball object (for racket sports the wrist joints connect to the ball, while for soccer the ankle and shoulder joints connect to the ball to approximate lower- and upper-body ball control); $\mathcal{E}^{\text{p--c}}_t$ links human foot joints to court corners (positional context). 
% The intra-person edges follow standard human-joint topology, and court nodes connect as a rectangle of 4 corners. Cross-entity links are sport-specific: for racket sports the wrist joints connect to the ball, while for soccer the ankle and shoulder joints connect to the ball to approximate lower- and upper-body ball control. 
All edges are undirected. See full connection rules in our code.

Our proposed \emph{Unified Multi-Entity Graph} flexibly integrates multiple players with objects and environmental cues. Compared to conventional skeleton-only graphs that ignore objects and context (resulting in loss of critical event information), our unified graph provides a richer and more holistic representation of the scene. By explicitly including these elements as nodes and edges, \model{} can represent fine-grained event cues that would otherwise be overlooked.

\subsubsection{Graph-based Spatio-Temporal Encoding} 
Given a sequence of unified multi-entity graphs $G=\{\mathcal{G}_t\}_{t=1}^T  \in \mathbb{R}^{T \times |\mathcal{V}|\times2}$ over a video clip of $T$ frames, 
we extract discriminative features via a specially designed spatio-temporal encoder, named \textbf{UMEG Block}, tailored for precise event spotting under few-shot conditions. The UMEG Block consists of a \textbf{spatial graph convolution network} that operates on the unified graph topology, combined with a novel, parameter-free \textbf{temporal multi-scale shift module} that enables efficient and effective temporal feature extraction. In the following, we describe the spatial GCN module and the temporal multi-scale shift mechanism.

\textbf{(1) Spatial GCN.} 
This spatial GCN layer updates node features by aggregating information from their neighbors in the graph. In matrix form, the output of one GCN layer is:
\begin{equation}
\mathcal{H}^{(\ell+1)} = \text{ReLU}(A^{(\ell)} \mathcal{H}^{(\ell)} W^{(\ell)}),
\end{equation}
where $A(\ell) \in \mathbb{R}^{|\mathcal{V}| \times |\mathcal{V}|}$ is the adjacency matrix employed
for spatial aggregation, $\mathcal{H}^{(\ell)} \in \mathbb{R}^{|\mathcal{V}| \times T \times d}$ symbolizes the hidden representation, and $W^{(\ell)} \in \mathbb{R}^{d \times d}$ is the weight matrix utilized for feature projection. Here, $|\mathcal{V}|$, $T$, and $d$ denote the number of nodes, frames, and hidden feature dimension, respectively. $\text{ReLU}(\cdot)$ is the ReLU activation function, and the superscript $\ell$ indicates the layer number. 

Unlike traditional GCNs in skeleton-based action recognition~\cite{duan2022pyskl,zhou2024blockgcn,liu2025revealing} that process each individual independently, our approach performs graph convolutions over the entire multi-entity graph. This allows for joint modeling of both human–human and human–entity interactions, capturing intra- and inter-entity relationships simultaneously. Therefore, it can exploit richer contextual cues essential for precise event spotting.

\textbf{(2) Temporal Multi-Scale Shift.}
To model temporal correlations, most skeleton-based action recognition models~\cite{yan2018spatial,shi2020skeleton,chen2021channel,duan2022pyskl,zhou2024blockgcn,liu2025revealing} apply multi-scale temporal convolution modules after each spatial GCN to capture frame-to-frame dynamics. However, in few-shot settings, such modules increase the number of trainable parameters, making training more difficult and prone to overfitting. To address this, we draw inspiration from TSM~\cite{lin2019tsm} and introduce a Temporal Multi-Scale Shift Module that efficiently captures temporal dynamics by shifting feature vectors along the temporal axis at multiple scales, without adding additional trainable parameters.
% Instead of parameterized temporal convolutions or RNNs (which reduce temporal resolution and increase parameters~\cite{yan2018spatial,shi2020skeleton,chen2021channel,duan2022pyskl,zhou2024blockgcn,liu2025revealing}), we adopt a \emph{Multi-Scale Temporal Shift} (MTS) inspired by TSM~\cite{lin2019tsm} to fuse temporal context with \emph{zero} extra parameters.
Let $\mathcal{H}_t^{(\ell)}\!\in\!\mathbb{R}^{|\mathcal{V}|\times d}$ be the input features at frame $t$ and layer $\ell$. We split channels into {static}, {forward-shift}, and {backward-shift} parts with fraction $\alpha$ (we use $\alpha=1/8$)
\begin{equation}
    \mathcal{H}_t^{(\ell)}=\big[\,\mathcal{H}_{t,\mathrm{static}}^{(\ell)} \,\|\, \mathcal{H}_{t,\mathrm{fwd}}^{(\ell)} \,\|\, \mathcal{H}_{t,\mathrm{bwd}}^{(\ell)}\,\big],
\end{equation}
where $\mathcal{H}_{t,\mathrm{static}}^{(\ell)} \in \mathbb{R}^{|\mathcal{V}|\times (1-2\alpha)d}$, and $\mathcal{H}_{t,\mathrm{fwd}}^{(\ell)},\, \mathcal{H}_{t,\mathrm{bwd}}^{(\ell)} \in \mathbb{R}^{|\mathcal{V}|\times \alpha d}$. $\,\|\,$ is the notation for (channel-wise) concatenation.
For temporal offsets $\Delta \in \{1,2,4\}$, define the \emph{bidirectional} shift operator
\begin{equation}
    \tilde{\mathcal{H}}_{t}^{(\ell,\Delta)} \;=\; 
    \big[\, \mathcal{H}_{t,\mathrm{static}}^{(\ell)} \;\|\; \mathcal{H}_{t-\Delta,\mathrm{fwd}}^{(\ell)} \;\|\; \mathcal{H}_{t+\Delta,\mathrm{bwd}}^{(\ell)} \,\big],
\end{equation}
with boundary handling by zero padding. 
% We also keep the unshifted stream $\tilde{\mathcal{H}}_{t}^{(\ell,0)} \!=\! \mathcal{H}_{t}^{(\ell)}$ for residual link.
Each shifted stream is then passed to the spatial GCN to update the node embeddings with temporal context
\begin{equation}
    Z_{t}^{(\ell,\Delta)} \;=\; \text{ReLU}(A^{(\ell)}\, \tilde{\mathcal{H}}_{t}^{(\ell,\Delta)}\, W_{}^{(\ell)}), \quad \Delta \in \{1, 2, 4\},
\end{equation}
where ${A}$ is the adjacency matrix, $W$ is the weight for spatial GCN, and $Z_{t}^{(\ell,\Delta)} \in \mathbb{R}^{|\mathcal{V}|\times d}$.
We then fuse multi-scale contexts by 
\begin{equation}
    U_{t}^{(\ell)} = \big\|_{\Delta \in \{1, 2, 4\}} \; \text{F}_1(Z_{t}^{(\ell,\Delta)}),
\end{equation}
where $\text{F}_1$ is a linear projection to down-scale the channel size to $\mathbb{R}^{|\mathcal{V}|\times  \lfloor d / |\Delta| \rfloor}$. The output is
\begin{equation}
    \mathcal{H}_{t}^{(\ell+1)} = \text{F}_{2}(\text{ReLU}(U_{t}^{(\ell))}) + \mathcal{H}_{t}^{(\ell)},
\end{equation}
where $\text{F}_2$ is a linear projection to scale the channel size to $\mathbb{R}^{|\mathcal{V}|\times d}$, and $\mathcal{H}_{t}^{(\ell)}$ is added via a residual connection.
This multi-scale temporal shift mechanism preserves frame-level resolution, expands the temporal receptive field across short, mid, and long ranges (via $\Delta \in \{1, 2, 4\}$), and introduces no additional trainable parameters.

% where $H_{t,\mathrm{static}}^{(l-1)}\!\in\!\mathbb{R}^{|\mathcal{V}|\times (1-\alpha)d},\;
% H_{t,\mathrm{shift}}^{(l-1)}\!\in\!\mathbb{R}^{|\mathcal{V}|\times \alpha d}$.
% For temporal offsets $\Delta\!\in\!\mathcal{D}=\{1,2,4\}$, define the (partial) shift operator
% \begin{equation}
%     \tilde{H}_{t}^{(l-1,\Delta)} \;=\; \big[\,H_{t,\mathrm{static}}^{(l-1)} \,\|\, H_{t-\Delta,\mathrm{shift}}^{(l-1)}\,\big],
% \end{equation}
% Each shifted stream is passed to the spatial GCN that shares the same graph structure and weights:
% \[
% Z_{t}^{(\ell,\Delta)} \;=\; \sigma\!\big(\tilde{D}^{-1/2}\tilde{A}\tilde{D}^{-1/2}\,\tilde{H}_{t}^{(\ell,\Delta)} W_{s}^{(\ell,\Delta)}\big).
% \]
% We then fuse multi-scale contexts by concatenation and $1{\times}1$ linear fusion:
% \[
% U_{t}^{(\ell)} \;=\; \big\|_{\Delta\in\mathcal{D}} Z_{t}^{(\ell,\Delta)},\qquad
% H_{t}^{(\ell+1)} \;=\; U_{t}^{(\ell)} W_{f}^{(\ell)}.
% \]
% This MTS block preserves frame-level resolution, broadens temporal receptive fields ($\Delta\!\in\!\{1,2,4\}$ captures short/mid/long cues), and avoids extra temporal parameters, yielding efficient spatio-temporal encoding tailored for frame-accurate PES.

\subsubsection{Multimodal Knowledge Distillation} While the graph-based model excels at leveraging structured keypoint information, its performance can degrade if the pose or object detections are unreliable (e.g. due to motion blur or occlusions). To enhance robustness, we propose a multimodal distillation framework where knowledge is transferred from the graph domain to the raw visual domain. We introduce a student network that operates directly on the RGB video frames and learn it under the guidance of the trained graph-based teacher (our \model{}). 

Let the frozen graph-based teacher encoder $\epsilon_{\text{tch}}$ map the unified graph sequence $G=\{\mathcal{G}_t\}_{t=1}^T$ to per-frame embeddings
$\mathbf{F}_{\text{tch}}=\epsilon_{\text{tch}}(G) \in \mathbb{R}^{T\times d}$.
The RGB student encoder $\epsilon_{\text{stu}}$ maps the video $X\in\mathbb{R}^{T\times H\times W\times 3}$ to
$\mathbf{F}_{\text{stu}}=\epsilon_{\text{stu}}(X)\in\mathbb{R}^{T\times d}$. $\epsilon_{\text{stu}}$ consists of a visual feature extractor followed by a bidirectional GRU to capture long-term dependencies and project the extracted features into the same dimensional space $d$ as the teacher encoder.
On all unlabeled clips $D_{\text{unlabel}}$, we minimize a feature-matching L2 loss
\begin{equation}
    \mathcal{L}_{\text{feat}} \;=\; \frac{1}{T}\sum_{t=1}^{T}\big\|\mathbf{F}_{\text{tch}}^{(t)}-\mathbf{F}_{\text{stu}}^{(t)}\big\|_2^2,
\end{equation}
The teacher encoder, initialized from the pretrained weights in the previous stage, remains frozen during distillation, while the student networks are trainable. A few-shot adaption is adopted to fine-tune the event localizer and classifier on $D_{label}$. At inference time, the student alone performs PES directly from RGB, inheriting robustness from the teacher’s structured graph supervision.

\subsection{Implementation Details}
In \model{}, we employ the unit GCN from~\cite{zhou2024blockgcn} as the spatial GCN layer for its strong performance and efficiency. The multi-scale temporal shift module applies shifts of $\Delta \in \{1, 2, 4\}$ frames. For distillation, we use VideoMAEv2~\cite{wang2023videomaev2}, an advanced transformer-based visual encoder, as the backbone for feature extraction (smaller variants also show consistent gains). Event localization and classification are performed by linear layers that output event probabilities and event types, respectively. The training protocol processes 96-frame sequences with a stride of 2. RGB frames are resized to 224 pixels in height, then randomly cropped to 224 $\times$ 224 to preserve essential visual information. Standard data augmentation (cropping, color jittering) enhances data diversity and model robustness during training but is omitted in testing.
% Each model performs dense per-frame classification to detect event types and precise timestamps. Given the severe class imbalance, where event frames constitute less than 3\% of the dataset, the loss weight for foreground classes is increased fivefold relative to background classes.  
Models are optimized with AdamW (initial learning rate 0.001 for \model{}, 0.0001 for VideoMAEv2 in distillation), using three linear warm-up steps followed by cosine annealing. Training is conducted on an RTX 4090 GPU. Further implementation details are provided in the Appendix.

\section{Experimental Results}
This section presents evaluation details and ablation studies.
% This section presents detailed information about the evaluation and ablation study.
% In this section, we present the datasets, quantitative comparisons, and ablation study of our approach. More details can be found in Appendix.

\begin{table*}[ht]
\caption{
Experimental results and ablation studies for fine-grained sports event detection across five datasets using \emph{100-clip} training data are reported with evaluation metrics F$1_{\text{evt}}$ and edit score. ``Params (M)'' refers to the number of model parameters. Top-performing results are \textbf{bolded}, while the best within each method category are \underline{underlined}. Our method \model{} 
%achieves 
outperforms all the assessed
state-of-the-art (SOTA) methods.
}
\centering
\small
% \resizebox{\linewidth}{!}{
\begin{tabular}{lrrrrrrrrrrc}
\toprule
 & \multicolumn{2}{c}{\rotatebox{0}{F$^3$Set-Tennis}} & \multicolumn{2}{c}{\rotatebox{0}{ShuttleSet}} & \multicolumn{2}{c}{\rotatebox{0}{FineGym-BB}} &\multicolumn{2}{c}{\rotatebox{0}{Figure Skating}} &\multicolumn{2}{c}{\rotatebox{0}{SoccerNet-BAS}} &\multirow{2}{*}{\rotatebox{0}{Params (M)}} \\
\cmidrule(r){2-3} \cmidrule(lr){4-5} \cmidrule(l){6-7} \cmidrule(l){8-9} \cmidrule(l){10-11} 
  & F$1_{\text{evt}}$ & Edit & F$1_{\text{evt}}$ & Edit & F$1_{\text{evt}}$ & Edit & F$1_{\text{evt}}$ & Edit & F$1_{\text{evt}}$ & Edit \\
\midrule
\multicolumn{5}{l}{\textit{(a) SOTA PES methods}} \vspace{1mm} \\
E2E-Spot$_\text{ 200MF}$~\cite{hong2022spotting} &2.6 &11.1 &35.6 &50.3 &40.1 &50.8 &29.9 &34.8 &15.3 &35.4 &\underline{4.5} \\%&39.6\\
E2E-Spot$_\text{ 800MF}$~\cite{hong2022spotting} &3.1 &13.3 &42.7 &54.6 &\underline{44.7} &\underline{53.1} &35.9 &42.7 &22.1 &\underline{43.1} & 12.6 \\%&151.4\\
T-DEED$_\text{ 200MF}$~\cite{xarles2024t}  &1.0 &4.2 &33.8 &41.7 &44.1 &48.4 &36.6 &33.5 &4.7 &8.8 &16.4 \\%&22.0\\
T-DEED$_\text{ 800MF}$~\cite{xarles2024t}  &1.5 &6.4 &30.7 &38.6 & 43.6 &48.3 &\underline{37.9} &\underline{40.6} &6.7 &14.3 &46.2 \\%&60.3 \\
% F$^3$ED & & &39.1 &49.9 & & & & & & &4.7 \\%&77.2\\
F$^3$ED~\cite{liu2025f} &\underline{3.9} &\underline{15.3} &\underline{44.1} &\underline{55.1} &43.8 &52.1 &36.1 &34.4 &\underline{22.7} &34.5 &4.7 \\%&77.2\\
\midrule
\multicolumn{5}{l}{\textit{(b) Skeleton-based PES variants}} \vspace{1mm} \\
MSG3D~\cite{liu2020disentangling} &5.2 &15.4 &39.7 &56.3 & 45.8 &50.1 &18.1 &32.8 &\underline{22.3} &39.5 &4.6 \\%&57.7 \\
AAGCN~\cite{shi2020skeleton} &4.9 &15.4 &44.7 &55.6 &42.8 &47.8 &24.8 &40.0 &22.1 &43.1 &5.4 \\%&45.5 \\
CTRGCN~\cite{chen2021channel}&5.5 &16.9 &40.3 &55.1 &\underline{46.6} &50.7 &28.2 &44.4 &\underline{22.3} &42.4 &3.1 \\%&18.2 \\
STGCN++~\cite{duan2022pyskl} &6.4 &18.0 &45.0 &57.7  &44.6 &50.0 &29.5 &46.8 &17.4 &42.7 &3.0 \\%&18.6\\
ProtoGCN~\cite{liu2025revealing} &6.6 &18.1 &46.8 &58.3 &41.4 &\underline{51.1} &25.3 &43.7 &17.8 &41.0 &5.6 \\%&49.2\\
BlockGCN~\cite{zhou2024blockgcn} &\underline{6.9} &\underline{18.3} &\underline{47.1} &\underline{59.4} &44.5 &49.1 &\underline{29.8} &\underline{48.2} &19.9 &\underline{43.3} &\underline{2.5} \\%&11.1\\
\midrule
\multicolumn{5}{l}{\textit{(c) Our approach}} \vspace{1mm} \\
\textbf{\model{}}  & \textbf{9.4} &\textbf{31.7} &\textbf{49.2} &\textbf{64.0} &\textbf{49.2} &\textbf{54.4} &\textbf{39.2} &\textbf{49.6} &\textbf{27.0} &\textbf{44.8} &\underline{\textbf{2.2}} \\%&11.3\\
$\textbf{\model{}}_\text{ distill}$ &\underline{\textbf{12.5}} &\underline{\textbf{40.7}} &\underline{\textbf{59.1}} &\underline{\textbf{69.0}} &\underline{\textbf{58.4}} &\underline{\textbf{61.2}} &\underline{\textbf{45.9}} &\underline{\textbf{56.2}} &\underline{\textbf{27.1}} &\underline{\textbf{50.8}} &67.8 \\%&2445.4 \\
\midrule
\emph{\textbf{Ablation studies}} \vspace{1mm}\\
% \midrule
\textit{(d)} $\text{pose} * N$ &5.6 &23.9 &47.4 &61.5 &\underline{49.2} &\underline{54.4} &\underline{39.2} &\underline{49.6} &20.7 &39.6&--\\
\;\;\;\;\;$\text{pose} * N + \text{court}$ &6.6 &26.1 &46.7 &61.5&--&--&--&--&--&--&--\\
\;\;\;\;\;$\text{pose} * N + \text{ball}$ &8.6 &30.2 &48.1 &62.5 &--&--&--&--&\underline{27.0} &\underline{44.8}&--\\
\;\;\;\;\;$\text{pose} * N + \text{ball} + \text{court}$ & \underline{9.4} &\underline{31.7} &\underline{49.2} &\underline{64.0} &--&--&--&-- &-- &-- &-- \\
\midrule
\textit{(e)} $\Delta \in \{1\}$ &8.8 &30.4 &46.5 &61.2 &39.0 &50.3 &26.2 &36.8 &21.1 &38.9 &--\\
\;\;\;\;\;$\Delta \in \{1, 2\}$ &\underline{9.6} &\underline{33.2} &47.4 &61.6 &42.9 &49.8 &32.1 &45.3 &23.1 &40.1 &--\\
\;\;\;\;\;$\Delta \in \{1, 2, 4\}$ & 9.4 &31.7 &\underline{49.2} &\underline{64.0} &\underline{49.2} &\underline{54.4} &\underline{39.2} &\underline{49.6} &\underline{27.0} &\underline{44.8} &--\\
\midrule
% \multicolumn{5}{l}{\textit{() Self-supervise methods}} \vspace{1mm} \\
% 2D-VPD~\cite{} & & & & & &\\
\textit{(f)} Self-supervise~\cite{chen2020simple} &3.0 &29.1 &50.2 &62.4 &54.5 &56.8 &34.6 &41.3 &26.0 & 42.9&--\\
\;\;\;\;\;\textbf{\model{}}$_\text{ distill}$ &\underline{{12.5}} &\underline{{40.7}} &\underline{{59.1}} &\underline{{69.0}} &\underline{{58.4}} &\underline{{61.2}} &\underline{{45.9}} &\underline{{56.2}} &\underline{{27.1}} &\underline{{50.8}} &-- \\
\midrule
\textit{(g)} E2E-Spot (full supervision) &44.6 &71.1 &71.2 &\underline{76.1} &\underline{72.9} &\underline{73.0} &58.0 &63.9 &\underline{46.2} &\underline{72.9} & --\\
\;\;\;\;\;\textbf{\model{}} (full supervision) &\underline{47.5} &\underline{71.2} &\underline{71.4} &\underline{76.1} &59.8 &64.8 &\underline{61.8} &\underline{71.8} &36.1 &55.7 &--\\
\bottomrule
\end{tabular}
% }
\label{tab:few-shot}
\end{table*}

\subsection{Datasets}
To evaluate the effectiveness of our method, we conduct experiments on several PES datasets, including racket sports F$^3$Set-Tennis~\cite{liu2025f} and ShuttleSet~\cite{wang2023shuttleset}, individual sports FineGym~\cite{shao2020finegym} and Figure Skating~\cite{hong2021video}, and team sports SoccerNet-BAS~\cite{cioppa2024soccernet2024challengesresults_short}. Detailed descriptions of these datasets are provided in the supplementary material.

% \subsection{Keypoint Acquisition}
We extract 2D poses using off-the-shelf pose estimators. For 2D pose estimation, we use HRNet~\cite{sun2019deep} to detect athletes' poses through top-down estimation. 
% We use only 12 of the 17 COCO~\cite{lin2014microsoft} keypoints (ignoring Nose, LEye, REye, LEar, and REar), and we apply the same joint normalization procedure as in~\cite{sun2020view}. 
For sports-specific object detection and tracking, we fine-tune the pretrained
YOLOv8~\cite{yolov8_ultralytics} model on corresponding public datasets from Roboflow~\cite{dwyer2025roboflow} to detect and track target players and sports balls in F3Set-Tennis, ShuttleSet, and SoccerNet-BAS. For two racket sports datasets, we also employ deployed models for court detection to identify the four corners. 
% For FineGym and Figure Skating, we only utilize keypoints from human poses. 
Please refer to Appendix for more details.

\subsection{Evaluation Metrics}  
Following Liu et al.~\cite{liu2025f}, we evaluate our method using two metrics that assess temporal precision and classification accuracy: Edit score and mean F1 with temporal tolerance.
\emph{(1) Edit Score}~\cite{lea2017temporal} measures the structural similarity between predicted and ground-truth event sequences using Levenshtein distance, accounting for missing, redundant, and misordered predictions. It is suitable for tasks requiring accurate event ordering and completeness.
\emph{(2) Mean F1 Score with Temporal Tolerance} evaluates both event classification and localization accuracy. A prediction is correct if it matches the event class and occurs within a temporal window of $\delta$ frames around the ground-truth timestamp. We report the average F1 across all event types, denoted as F$1_{\text{evt}}$. Unless otherwise specified, we use a strict tolerance of $\delta = 1$ frame; for SoccerNet-BAS, we follow prior work~\cite{cioppa2024soccernet2024challengesresults_short} and use $\delta = 1$ second.

% We follow~\cite{liu2025f} and evaluate our method using two metrics - Edit score and mean F1 with temporal tolerance - that assess both temporal precision and classification accuracy.  
% \emph{Edit Score}~\cite{lea2017temporal} quantifies the structural similarity between predicted and ground truth event sequences using Levenshtein distance, capturing errors such as missing, extra, or misordered events. This metric is particularly useful for evaluating models where sequence order and completeness are critical~\cite{he2024vistec}.  
% \emph{Mean F1 Score with Temporal Tolerance}~\cite{hong2022spotting,he2024vistec} measures classification and temporal localization accuracy. A prediction is considered correct if its timestamp falls within a strict tolerance range $\delta$ (e.g., $\pm1$ frame) and the event class is correctly identified, ensuring precise temporal spotting alongside accurate classification. We report F$1_{\text{evt}}$, which is the average F1 score across all event types. We set a tight tolerance with $\delta=1$ frame except for SoocerNet-BAS, where $\delta=1$ second is employed~\cite{cioppa2024soccernet2024challengesresults_short}.
% two variants: F$1_{\text{lcl}}$, which evaluates the module’s precision in identifying event moments, and F$1_{\text{evt}}$, the average F1 score across all event types.  

\subsection{Baselines}  \label{baseline}
We compare our approach with SOTA PES methods that operate on RGB inputs and support end-to-end training, including E2E-Spot~\cite{hong2022spotting} and T-DEED~\cite{xarles2024t}, each evaluated with RegNet-Y~\cite{radosavovic2020designing} 200MF and 800MF backbones, as well as F$^3$ED~\cite{liu2025f}. Furthermore, we construct skeleton-based PES variants of these baselines by replacing their visual encoders with existing skeleton-based GCN architectures. These variants take 2D human poses as input and employ F$^3$ED's head module for event spotting. Specifically, we adopt graph-based models including MSG3D~\cite{liu2020disentangling}, AAGCN~\cite{shi2020skeleton}, CTRGCN~\cite{chen2021channel}, STGCN++~\cite{duan2022pyskl}, BlockGCN~\cite{zhou2024blockgcn}, and ProtoGCN~\cite{liu2025revealing}. Per-frame representations are computed by aggregating individual skeleton features via averaging across all detected persons.

\subsection{Few-Shot Setting} 
We define the few-shot scenario as training with a limited number of annotated clips ($k$-clip), following~\cite{hong2021video}. In F$^3$Set-Tennis and ShuttleSet, each clip represents a rally spanning several seconds and comprising a sequence of shots. In FineGym-BB and Figure Skating, each clip corresponds to a routine with a series of technical and artistic movements. In SoccerNet-BAS, each clip captures a phase of play featuring actions such as pass, drive, and shoot.

Unlike the traditional $k$-shot approach, which segments videos into independent samples containing single events with backgrounds~\cite{yang2020localizing,nag2021few}, our $k$-clip strategy offers a more practical and domain-adapted setting. \emph{First}, due to the large number of event types in sports, many of which are rare, sampling $k$ instances per type is often impractical; annotating a small set of clips per sport domain is more efficient and scalable. \emph{Second}, sports actions are typically brief (1–2 frames) and occur rapidly in succession, thus isolated events lack essential temporal context. \emph{Third}, consecutive events frequently exhibit strong dependencies~\cite{hong2022spotting,liu2025f}; therefore, training on $k$-clips rather than $k$-shots enables the model to better capture long-term event relationships.

\begin{figure*}[ht]
\centering
    \includegraphics[width=\linewidth]{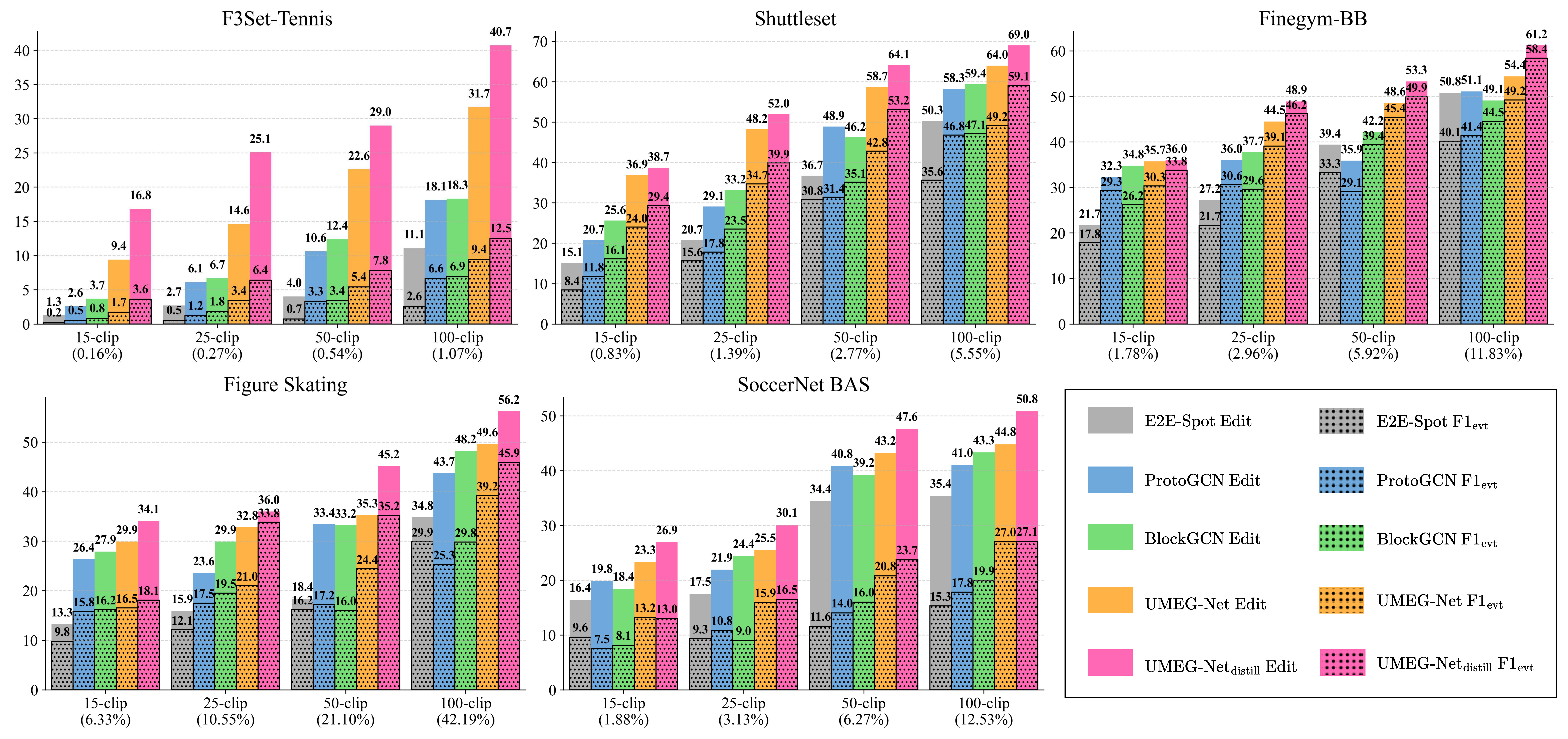}
\caption{F1$_\text{evt}$ and Edit scores under few-shot ($k$-clip) training. Percentages indicate the fraction of the full training set.}
% \caption{F1$_\text{evt}$ and Edit scores across five datasets under few-shot ($k$-clip) training. Percentages indicate the fraction of the full training set. \model{} and \model{}$_\text{ distill}$ outperforms all baselines across all datasets.}
\label{fig:k-clip}
\end{figure*}

\subsection{Result Analysis with Few-Shot Supervision}
We evaluate performance in the few-shot setting across all datasets. For a fair comparison, all methods are trained on the same $k$-clip samples, with five random splits per dataset, and the results are averaged. We test $k \in \{15, 25, 50, 100\}$ across SOTA PES methods, skeleton-based PES variants, and our methods.
% \model{} and $\model{}_{\text{ distill}}$. 
Table~\ref{tab:few-shot} presents the results for the 100-clip setting.
From the table, we observe that \model{} consistently achieves strong performance across all datasets. 

\textbf{(1) Comparison with SOTA PES methods.} \model{} achieves substantial improvements over these methods as shown in Table~\ref{tab:few-shot}(a). Specifically, it improves F$1_{\text{evt}}$ by 1.3\% to 5.5\% and Edit score by 1.3\% to 16.4\% over the best-performing PES baselines across five datasets, respectively. Despite strong performance on large-scale labeled data, these PES methods generally struggle in few-shot settings, highlighting the limitation of relying solely on RGB inputs.

% \emph{First}, compared to existing PES methods that take RGB images as inputs, \model{} achieves substantial improvements in both evaluation metrics. Specifically, it improves F$1_{\text{evt}}$ by 3.1\% to 5.5\% and Edit score by 1.3\% to 16.4\% over the best-performing PES baselines across the five datasets, respectively. Despite strong performance on large-scale labeled data, these PES methods struggle in few-shot settings.
% Despite strong performance on large-scale labeled data, these PES methods generally struggle in few-shot settings, highlighting the importance of our approach’s effectiveness in low-data regimes.

\textbf{(2) Compare with skeleton-based PES variants.} \model{} shows noticeable advantages over skeleton-based variants in Table~\ref{tab:few-shot}(b), especially on datasets containing fine-grained events such as F$^3$Set-Tennis and ShuttleSet. For instance, it outperforms the best-performing variant BlockGCN by +2.5\% F$1_{\text{evt}}$ and +13.4\% Edit on F$^3$Set-Tennis, and +2.1\% F$1_{\text{evt}}$ and +4.6\% Edit on ShuttleSet. This advantage stems from \model{}’s incorporation of object and environmental cues, as well as its compact architecture, with the fewest parameters (2.2M) among all baselines.

% \emph{Second}, \model{} shows noticeable advantages over skeleton-based PES variants, especially on datasets containing fine-grained contextual events such as F$^3$Set-Tennis and ShuttleSet. For instance, it surpasses the best-performing variant BlockGCN by +2.5\% F$1_{\text{evt}}$ and +13.4\% Edit on F$^3$Set-Tennis, and by +2.1\% F$1_{\text{evt}}$ and +5.0\% Edit on ShuttleSet. 

% This advantage stems from \model{}’s incorporation of object and environmental cues, as well as our compact architecture design, with the fewest parameters (2.2M) among all baselines.
% This advantage mainly stems from \model{}’s effective integration of object and environmental cues, coupled with a compact architecture having the fewest parameters (2.2M) among all baselines. 
% These findings underline the importance of multi-entity graph and .

\textbf{(3) Multimodal distillation.} $\model{}_{\text{ distill}}$ further enhances performance and robustness as shown in Table~\ref{tab:few-shot}(c). Compared to \model{}, it improves on average 5.8\% in F$1{\text{evt}}$ and 6.7\% in Edit score. These improvements demonstrate the effectiveness of multimodal distillation in capturing complementary visual representations from RGB inputs.

% \emph{Third}, $\model{}_{\text{ distill}}$ further enhances performance and robustness. Compared to \model{}, it achieves an average improvement of +5.8\% in F$1{\text{evt}}$ and +6.7\% in Edit score. These improvements demonstrate the effectiveness of multimodal distillation in capturing complementary visual representations from RGB inputs.
\textbf{(4) Various $k$-clip settings.}
Figure~\ref{fig:k-clip} illustrates performance trends for representative PES baselines and our proposed models across \emph{various $k$-clip samples}. The figure clearly indicates that \model{} consistently outperforms existing methods under all supervision levels, and $\model{}_{\text{distill}}$ provided further improvement. These results confirm the robustness and effectiveness of our approach across diverse few-shot scenarios.

\subsection{Ablation Studies}
We conduct ablation studies to evaluate the effectiveness of our proposed framework in few-shot settings, primarily focusing on the 100-clip setup unless otherwise specified.

% \begin{table}
% \caption{
% Ablation studies for PES using \emph{100-clip} few-shot training data are reported with evaluation metrics F$1_{\text{evt}}$ and edit score.
% }
% \centering
% \small
% \resizebox{\linewidth}{!}{
% \begin{tabular}{lcccccccccccccc}
% \toprule
%  & \multicolumn{2}{c}{\rotatebox{0}{F$^3$Set-T}} & \multicolumn{2}{c}{\rotatebox{0}{ShuttleSet}} &\multicolumn{2}{c}{\rotatebox{0}{Gym}} &\multicolumn{2}{c}{\rotatebox{0}{FS}} &\multicolumn{2}{c}{\rotatebox{0}{SoccerNet-BAS}} \\
% \cmidrule(r){2-3} \cmidrule(lr){4-5} \cmidrule(l){6-7} \cmidrule(l){8-9} \cmidrule(l){10-11} 
%   & F$1_{\text{evt}}$ & Edit & F$1_{\text{evt}}$ & Edit & F$1_{\text{evt}}$ & Edit & F$1_{\text{evt}}$ & Edit & F$1_{\text{evt}}$ & Edit  \\
% \midrule
% \multicolumn{5}{l}{\textit{() Our approach}} \vspace{1mm} \\
% \textbf{\model{}}  & 9.4 &31.7 &49.2 &64.0  &27.0 &44.8\\ %&28.4 &49.1
% \textbf{\model{}} w distill &12.5 &40.7 &59.1 &69.0  &27.1 &50.8 \\
% \midrule
% $\text{pose} * P$ &5.6 &23.9 &47.4 &61.5 &20.7 &39.6\\
% $\text{pose} * P + \text{court}$ &6.6 &26.1 &46.7 &61.5&--&--\\
% $\text{pose} * P + \text{ball}$ &8.6 &30.2 &48.1 &62.5 &27.0 &44.8\\

% \midrule
% Contrastive~\cite{} & & &50.2 &62.4 \\

% % \textbf{\model{}} w distill &12.5 &40.7 &50.9 &69.0 &60.4 &61.2 &45.9 &56.2 &27.1 &50.8 \\
% \bottomrule
% \end{tabular}
% }
% \label{tab:ablation}
% \end{table}

\subsubsection{Effect of Graph Entities}
To assess the impact of different entity types, we analyze graph structures incorporating various object configurations, as shown in Table~\ref{tab:few-shot}(d). 
Specifically, ``$\text{pose} \times N$'', ``$\text{ball}$'', and ``$\text{court}$'' represent $N \times K$ human joint keypoints, a single keypoint for the ball, and four keypoints for court corners, respectively. 
The results indicate that incorporating ball or court information improves performance over using pose keypoints alone, while combining all entity types yields the best performance.

% \subsubsection{Effect of Graph Entities}
% To investigate the effect of entity information, we examined graph structures incorporating objects in various configurations, as presented in Table~\ref{tab:few-shot}(d), where $\text{pose} * N$, $\text{ball}$, $\text{court}$ represent $N * K$ joint keypoints from $N$ persons, 1 keypoint from the ball, and 4 keypoints for the court corners, respectively. As shown in the table,  the inclusion of ball objects or court information can enhance the performance compared to use soly pose keypoints. And incorporating all information gives the best performance.

% pose*M+{ball/+net} consists of MV + v nodes, including
% M people and v object points (ball/+basketball net) connected to the hands of
% each actor. (pose-{ball/+net})*M introduces individual object nodes for each
% person, which are initialized with the same data. Placing the object keypoints
% separately from the multi-person skeletons leads to significantly lower accuracy
% compared to having individual non-shared object keypoints. Additionally, the
% inclusion of extra objects, such as the basketball net, further enhances recognition accuracy, indicating the effectiveness and generalizability of the proposed
% panoramic graph.

\subsubsection{Temporal Module Configuration} 
% In \model{}, we adopted a multi-scale temporal shift module to incorporate temporal features. 
We examine the choice of $\Delta$ in \model{} to see how it affects overall performance.
We evaluate the effectiveness of different scales in temporal module by shifting $\Delta \in \{1\}$, $\Delta \in \{1, 2\}$, and compare with our default setting of $\Delta \in \{1, 2, 4\}$. As presented in Table~\ref{tab:few-shot}(e), the performance drops with less temporal shift scales. However, further increasing the scales does not necessarily increase the performance.

\subsubsection{Self-Supervise vs. \model{}$_\text{ distill}$} To validate the effectiveness of our multimodal distillation from \model{} to visual encoders, we compare it with a visual-based self-supervise alternative approach, where the visual models is pretrained on unlabeled domain data using contrastive learning~\cite{chen2020simple} and fine-tuned on k-clip labeled samples. As shown in Table~\ref{tab:few-shot}(f), although the contrastive learning approach show notable improvement compared to visual-based method trained only on labeled data, our \model{}$_\text{ distill}$ substantially outperforms it.

\subsubsection{Full Supervision Performance} We also compare our \model{} with existing RGB-based PES methods (E2E-Spot) under full supervision, where all of labeled clips in
the training set are available. As shown in Table~\ref{tab:few-shot}(g), our model is competitive even in the high-data regime. It shows better performance in 3 out of 5 datasets. This shows that \model{} is not limited to few-shot regimes.

\section{Conclusion}
In this paper, we introduced the task of few-shot Precise Event Spotting (PES), addressing the critical challenge of frame-accurate event detection in sports videos with limited labeled data. To tackle this, we proposed a unified multi-entity graph representation that flexibly integrates human skeletons, sport-specific object keypoints, and contextual landmarks, enabling richer modeling of complex interactions within diverse sports scenes. Leveraging this representation, we developed \model{}, a lightweight yet powerful graph-based architecture utilizing spatial graph convolutions combined with a parameter-free multi-scale temporal shift mechanism, enhanced further by multimodal knowledge distillation. Extensive experiments across five fine-grained sports event datasets demonstrated that our approach significantly outperforms existing methods in few-shot scenarios. 
% and remains competitive under full supervision. 
These results confirm the effectiveness of structured graph representations and multimodal learning strategies in addressing annotation scarcity and generalization challenges in precise event spotting. Future work can extend our approach to handle events with weak or non-entity cues, improving robustness beyond entity-driven scenarios.
% However, our method is most effective for events that strongly involve detectable entities (e.g., human–object interactions in sports) but may be less suitable for domains with weak or non-entity-centered cues, which we leave for future investigation.

\section{Acknowledgments}
This research is supported by the National Research Foundation Singapore under its AI Singapore Programme (Award Number: AISG3-RP-2022-030).  Any opinions, findings, and conclusions expressed in this material are those of the author(s) and do not reflect the views of funding bodies.

% \clearpage

\bibliography{aaai2026}

\clearpage

\appendix
\section*{Appendix}
\section{Implementation Details} \label{implementation_details}
% This section provides additional implementation details for \model{} described in Section~\ref{approach}. Our code is publicly available at the anonymous link: \url{https://anonymous.4open.science/r/MD-FED/}.

% \paragraph{Pre-processing} In the pre-processing step, we extract 2D poses and optical flow using off-the-shelf estimators. For 2D pose estimation, we use HRNet~\cite{sun2019deep} to detect athletes' poses through top-down estimation, utilizing all 17 COCO keypoints~\cite{lin2014microsoft}. The 2D joint positions are normalized by rescaling and centering, following~\cite{sun2020view}. For optical flow extraction, we apply RAFT~\cite{teed2020raft} to compute flow between consecutive frames. The final flow is obtained by subtracting the median, clipping values to ±20 pixels, and quantizing to 8 bits.  

\subsubsection{Architecture} 
For RGB feature extraction, we employ VideoMAEv2~\cite{wang2023videomaev2} as the backbone, chosen for its superior classification performance. The encoder is pretrained on Kinetics-710~\cite{kay2017kinetics} and subsequently finetuned on the target sports datasets. VideoMAEv2 divides long video clips into slices and processes each slice independently, with a default slice length of 16 frames. However, in precise event spotting (PES), frame-level granularity is essential due to the short duration of sports events. Therefore, we set the slice length to 2 frames. Each slice, shaped $H \times W \times 2$ ($H=224$, $W=224$), is first passed through a 3D convolutional layer, resulting in $16 \times 16$ feature patches per slice. Each patch encodes the spatial features of its corresponding region. The sequence of patches is then concatenated and processed by a transformer-based network, which models relationships among spatial regions within the slice. The temporal features of the video are further modeled using the spatial features from all slices. After spatial modeling, we obtain features of shape $T/2 \times 196 \times d''$, which are temporally upsampled by a factor of 2 to $T \times 196 \times d''$. These features are subsequently fed into a temporal block to capture long-term dependencies and are used by the event localizer and classifier to predict frame-wise events.

\subsubsection{Training} 
For \model{}, we conduct dense, per-frame classification to detect event types and precise timestamps. Due to severe class imbalance—event frames account for less than 3\% of the data—the loss for foreground classes is increased fivefold to mitigate this disparity. The models are optimized using AdamW with a cosine annealing learning rate schedule, and training is performed on an RTX 4090 GPU. \model{} is trained end-to-end on the limited labeled data for 50 epochs (30 epochs for ShuttleSet due to its smaller size), with an initial learning rate of 0.001 and three linear warm-up steps before cosine decay. Model selection is based on the best validation performance.

For the multimodal distillation stage, the trained \model{} serves as the teacher to ``distill'' knowledge to the video encoder using a large set of unlabeled clips under weak supervision. \model{}$_{\text{distill}}$ is trained on both labeled and unlabeled data for 50 epochs (30 for ShuttleSet) with an initial learning rate of 0.0001. Following distillation, the video encoder parameters are frozen, and only the temporal block, event localizer, and classifier are finetuned on the labeled data to adapt to the PES task. This finetuning is conducted for 10 epochs with a learning rate of 0.001.

\section{Dataset Details}
This section provides additional information on the datasets used in our experiments, as well as details on the extraction of relevant keypoints.

\paragraph{F$^3$Set-Tennis} is a large-scale PES dataset proposed by Liu et. al.~\cite{liu2025f}. It consists of 11,584 clips from 114 professional tennis matches, with each clip containing 1 to 34 shots. Each shot is annotated with the exact frame of racket-ball contact and its corresponding event type. We focus on all 8 sub-classes and 1,108 event types. Detailed statistics are shown in Table\ref{tab:f3set-tennis}.

For this dataset, we extract keypoint information comprising players’ 2D poses, ball positions, and court corners. The near-side player is identified using a detect-and-track algorithm~\cite{girdhar2018detect}, selecting the trajectory closest to the bottom of the video frame. To distinguish the far-side player from line judges, we apply YOLOv8~\cite{yolov8_ultralytics} to detect individuals above the net and assign as the far-side player the lowest detected person within three meters of the court boundary. Both players’ 2D poses and bounding boxes are estimated with YOLOv8. To account for camera motion, we detect court lines using the method of Jiang et al.~\cite{jiang2023court} and compute a homography to map the image plane to the court’s ground plane. Assuming the player’s feet are on the ground, we localize each player by projecting the bottom center of their bounding box onto the court. Ball positions in each frame are detected via~\cite{jiang2024tracking}.

\begin{table}[t]
\caption{Distribution of elements across sub-classes in the F$^3$Set-Tennis.}
\centering
\begin{tabular}{@{}llrcccccccccccccc@{}}
\toprule
 Sub-Class & Element & Count & Proportion (\%)\\
\midrule
\multirow{2}{*}{$sc_1$} & {near} & 21,467 & 50.1\% \\
                        & {far} & 21,362 & 49.9\%\\ 
\midrule
\multirow{3}{*}{$sc_2$} & {deuce} & 14,474 & 33.8\% \\
                        & {ad} & 16,310 & 38.1\%\\
                        & {middle} & 12,045 & 28.1\%\\
\midrule
\multirow{2}{*}{$sc_3$} & {forehand} & 27,802 & 64.9\%\\
                        & {backhand} & 15,027 & 35.1\%\\ 
\midrule
\multirow{3}{*}{$sc_4$} & {serve} & 11,584 & 27.0\% \\
                        & {return} & 8,216 & 19.2\% \\
                        & {stroke} & 23,029 & 53.8\% \\
\midrule
\multirow{8}{*}{$sc_5$} & {T} & 4,428 & 10.3\%\\
                        & {Body} & 2,241 &  5.2\%\\
                        & {Wide} & 4,915 &  11.5\%\\
                        & {cross-court} & 11,933 & 27.9\% \\
                        & {down the line} & 3,521 & 8.2\% \\
                        & {down the middle} & 11,040 & 25.8\% \\
                        & {inside-in} & 608 & 1.4\% \\
                        & {inside-out} & 4,143 & 9.7\% \\
\midrule
\multirow{6}{*}{$sc_6$} & {ground stroke} & 38,287 & 89.4\% \\
                        & {slice} & 3,358 & 7.8\% \\
                        & {volley} & 497 & 1.2\% \\
                        & {lob} & 334 & 0.8\% \\
                        & {drop} & 236 & 0.5\% \\
                        & {smash} & 117 & 0.3\% \\
\midrule
\multirow{2}{*}{$sc_7$} & {approach} & 964 & 2.3\% \\
                        & {non-approach} & 41,865 & 97.7\% \\
\midrule
\multirow{4}{*}{$sc_8$} & {in-bound} & 31,245 & 73.0\% \\
                        & {winner} & 3,734 & 8.7\% \\
                        & {forced error} & 2,808 & 6.5\% \\
                        & {unforced error} & 5,042 & 11.8\% \\
\bottomrule
\end{tabular}
\label{tab:f3set-tennis}
\end{table}

\paragraph{ShuttleSet} is a publicly available badminton singles dataset featuring stroke-level annotations~\cite{wang2023shuttleset}. It comprises 104 sets, 3,685 rallies, and 36,492 strokes across 44 matches played between 2018 and 2021 by 27 top-ranking men's and women's singles players. While originally intended for tactical analysis, the dataset also provides detailed stroke types, precise stroke timestamps, and corresponding videos, making it suitable for the PES task. For our study, we construct the ShuttleSet dataset, which includes 3,685 clips (rallies), with an average clip length of 10.9 seconds and 10.5 shots per rally. Each shot is annotated with the exact racket–shuttle contact frame and its event type (36 categories). Detailed statistics are shown in Table\ref{tab:shuttleset}.

We extract keypoint information in ShuttleSet similar to F$^3$Set-Tennis, including players’ 2D poses, shuttlecock positions, and court corners. Court corners are detected using the approach described in~\cite{liu2022monotrack}, and shuttlecock tracking is performed with TrackNetV3~\cite{chen2023tracknetv3}. For player tracking, we use YOLOv8~\cite{yolov8_ultralytics} for human detection, filtering out any detected players whose feet are not within the court, and distinguish near and far players based on their distance to the camera. The 2D poses of detected players are estimated using a pre-trained HRNet~\cite{sun2019deep}.

\paragraph{FineGym} is a gymnastics dataset designed for fine-grained action understanding~\cite{shao2020finegym}. The original annotations specify the start and end times of each action, which we treat as discrete events (e.g., ``balance beam dismount start'' and ``balance beam dismount end'') in line with~\cite{hong2022spotting}. Our study focuses on the balance beam subset, referred to as FineGym-BB, which consists of 1,112 routines from 142 matches, with an average clip duration of 92 seconds and 24.8 events per clip. Event distributions are detailed in Table~\ref{tab:finegym}. For FineGym-BB, we extract 2D human poses of the single gymnastic athlete in each video clip using~\cite{sun2019deep}.

% Each annotation indicates action start and end boundaries, which are treated as distinct events such as ``balance beam dismount start'' and ``balance beam dismount end''. In constructing the FineGym PES dataset, we disregard the original splits—designed for action recognition and containing overlapping videos—and instead adopt a 3:1:1 train/validation/test split. To address variable source frame rates (25–60 FPS), all videos at 50 and 60 FPS are resampled to 25 and 30 FPS, respectively. Manual inspection revealed that start-frame annotations are more visually consistent than end-frame labels; for example, end frames for jumps frequently occur several frames after landing. Consequently, we also report results for a start-event-only subset, FineGym-Start, containing only action start annotations. FineGym-BB is the balance beam subset of the FineGym PES dataset, as described above.

% \paragraph{FineGym-BB}~\cite{shao2020finegym} focuses on the balance beam subset of the FineGym dataset. The original annotations specify action start and end times, which we treat as discrete events (e.g., ``balance beam dismount start'' and ``balance beam dismount end''). The dataset contains 1,112 routines from 142 matches, with an average clip duration of 92 seconds and 24.8 events per clip. Event distributions are shown in Table~\ref{tab:finegym}.

\paragraph{Figure Skating} includes 11 videos covering 371 short program performances. Following~\cite{hong2022spotting}, the dataset defines 20 event types corresponding to the take-off and landing frames of 10 jump and flying spin classes (e.g., ``axel take-off,'' ``flip landing''). Each program lasts 170.7 seconds on average, containing approximately 10 events per performance. Event distributions are detailed in Table~\ref{tab:fs}. Similarly, we only extract 2D human poses of the single figure skating athlete in each video clip using~\cite{sun2019deep}.

\paragraph{SoccerNet Ball Action Spotting (BAS)} focuses on identifying both the timing and type of ball-related actions across 12 classes~\cite{cioppa2024soccernet2024challengesresults_short}, with each action annotated by a single timestamp. The classes include Pass, Drive, Header, High Pass, Out, Cross, Throw In, Shot, Ball Player Block, Player Successful Tackle, Free Kick, and Goal. The original dataset consists of seven untrimmed broadcast videos of full English Football League matches, which contain many irrelevant scenes beyond key ball action events. To address this, we segment the videos to retain only segments containing ball action events.

For the resulting SoccerNet BAS clips, we extract 2D human poses and soccer ball positions. Soccer ball detection is performed using a YOLOv8 model pre-trained on an annotated Roboflow dataset~\cite{soccernet-uoejr_dataset}, while human 2D poses are estimated in a top-down manner using HRNet~\cite{sun2019deep}.

\begin{table}
% \small
\centering
\caption{Distribution of event types in ShuttleSet.}
\begin{tabular}{lr}
\toprule
{Event type}        & {Count}                     \\ 
\midrule
far-end player back-court-drive & 169 \\
far-end player clear & 905 \\
far-end player cross-court-net-shot & 479 \\
far-end player defensive-return-drive & 99 \\
far-end player defensive-return-lob & 82 \\
far-end player drive & 233 \\
far-end player driven-flight & 18 \\
far-end player drop & 711 \\
far-end player lob & 1,710 \\
far-end player long-service & 159 \\
far-end player net-shot & 2,157 \\
far-end player passive-drop & 455 \\
far-end player push & 1,014 \\
far-end player return-net & 1,180 \\
far-end player rush & 190 \\
far-end player short-service & 946 \\
far-end player smash & 937 \\
far-end player wrist-smash & 586 \\
near-end player back-court-drive & 153 \\
near-end player clear & 925 \\
near-end player cross-court-net-shot & 467 \\
near-end player defensive-return-drive & 173 \\
near-end player defensive-return-lob & 102 \\
near-end player drive & 251 \\
near-end player driven-flight & 19 \\
near-end player drop & 681 \\
near-end player lob & 1,860 \\
near-end player long-service & 201 \\
near-end player net-shot & 2,091 \\
near-end player passive-drop & 434 \\
near-end player push & 990 \\
near-end player return-net & 1,277 \\
near-end player rush & 133 \\
near-end player short-service & 978 \\
near-end player smash & 773 \\
near-end player wrist-smash & 534 \\
\midrule
Total & 24,072 \\
\bottomrule
\end{tabular}
\label{tab:shuttleset}
\end{table}
% \normalsize

\begin{table}
% \small
\centering
\caption{Distribution of event types in FineGym-BB.}
\begin{tabular}{lr}
\toprule
{Event type}        & {Count}                     \\ 
\midrule
BB-dismounts start & 1,112 \\
BB-dismounts end & 1,112 \\
BB-flight-handspring start & 2,714 \\
BB-flight-handspring end & 2,714 \\
BB-flight-salto start & 4,123 \\
BB-flight-salto end & 4,123 \\
BB-leap-jump-hop start & 4,602 \\
BB-leap-jump-hop end & 4,602 \\
BB-turns start & 1,265 \\
BB-turns end & 1,265 \\
\midrule
Total & 27,632 \\
\bottomrule
\end{tabular}
\label{tab:finegym}
\end{table}
% \normalsize

\begin{table}
% \small
\centering
\caption{Distribution of event types in Figure Skating.}
\begin{tabular}{lr}
\toprule
{Event type}        & {Count}                     \\ 
\midrule
axel takeoff & 371 \\
axel landing & 371 \\
flip takeoff & 184 \\
flip landing & 184 \\
flying-camel takeoff & 216 \\
flying-camel landing & 216 \\
flying-sit takeoff & 151 \\
flying-sit landing & 151 \\
flying-upright takeoff & 6 \\
flying-upright landing & 6 \\
loop takeoff & 94 \\
loop landing & 94 \\
lutz takeoff & 248 \\
lutz landing & 248 \\
salchow takeoff & 61 \\
salchow landing & 61 \\
toe-loop takeoff & 504 \\
toe-loop landing & 504 \\
\midrule
Total & 3,670 \\
\bottomrule
\end{tabular}
\label{tab:fs}
\end{table}
% \normalsize

\begin{table}
% \small
\centering
\caption{Distribution of event types in Figure Skating.}
\begin{tabular}{lr}
\toprule
{Event type}        & {Count}                     \\ 
\midrule
Pass & 4,955 \\
Drive & 4,274 \\
Head & 707 \\
High Pass & 756 \\
Out & 550 \\
Cross & 260 \\
Throw In & 359 \\
Shot & 168 \\
Ball Player Block & 222 \\
Player Successful Tackle & 74 \\
Free Kick & 19 \\
Goal & 13 \\
\midrule
Total & 12,357 \\
\bottomrule
\end{tabular}
\label{tab:soccernet}
\end{table}

\section{Baseline Methods}
This section provides additional implementation details for the baseline methods discussed in \emph{Experimental Results}.

\paragraph{E2E-Spot~\cite{hong2021video}} E2E‑Spot is an end‑to‑end deep learning framework for temporally precise spotting of fine‑grained events in video, defined as predicting the exact frame (within 1–2 frames tolerance) when an event occurs. It integrates a per-frame CNN (e.g. RegNet‑Y~\cite{radosavovic2020designing} with GSM~\cite{sudhakaran2020gate}) to efficiently process hundreds of consecutive frames and a lightweight bidirectional GRU~\cite{dey2017gate} to model long-term temporal context. Unlike prior two-stage methods (feature extraction followed by separate head training), E2E‑Spot jointly learns spatial–temporal representations directly from raw pixels under end‑to‑end supervision, enabling both fine-grained motion sensitivity and global temporal reasoning.

\paragraph{TDEED~\cite{xarles2024t}} is a PES model that improves upon baseline methods (i.e., E2E‑Spot) by explicitly enhancing frame-level discriminability and preserving high temporal resolution across multiple scales. The architecture integrates a feature extractor (e.g., RegNet‑Y with local temporal modules) to produce per-frame tokens, followed by a temporal encoder‑decoder that downsamples and then restores the original temporal resolution via skip‑connections. Within this module, Scalable‑Granularity Perception (SGP) layers, and in particular the SGP‑Mixer variants, boost token discriminability by reducing similarity among adjacent frames. The combined architecture allows T‑DEED to model both local and global temporal context while retaining precision.

\paragraph{F$^3$ED~\cite{liu2025f}} is an end-to-end deep learning framework devised to detect and timestamp sequences of fast, frequent, fine‑grained (F$^3$) events, particularly suited to sports domains such as tennis, where events are brief, rapidly occurring, and highly detailed. The model begins by encoding consecutive video frames with a visual backbone, yielding dense per-frame features. A contextual refinement module then processes this sequence, producing precise temporal predictions for event sequences while preserving positional accuracy at the frame level. 

\paragraph{MSG3D~\cite{liu2020disentangling}} extends STGCN~\cite{yan2018spatial} by introducing multi-scale graph convolutions, enabling hierarchical feature extraction across different joint neighborhoods. It learns both local and global skeletal dependencies by applying multiple graph convolutions at varying scales. MSG3D enhances pose-based action recognition by capturing complex motion structures across multiple granularities.

\paragraph{AAGCN~\cite{shi2020skeleton}} incorporates adaptive adjacency learning to dynamically refine graph connections based on input features. Unlike STGCN, which uses a predefined skeleton topology, AAGCN learns data-driven spatial dependencies, allowing more flexible representation of human motion. This improves robustness to variations in pose estimation noise and enhances action recognition performance.

\paragraph{CTRGCN~\cite{chen2021channel}} introduces channel-wise topology refinement by modeling multi-channel dependencies within the graph structure. Instead of treating each joint's features independently, CTRGCN applies cross-channel interactions to capture co-occurring motion patterns. This improves feature expressiveness and enables better generalization in skeleton-based action detection.

\paragraph{STGCN++~\cite{duan2022pyskl}} is an optimized version of STGCN that enhances efficiency and representation capacity. It refines spatial-temporal graph convolutions by introducing lightweight architectural modifications, improving performance while maintaining a compact parameter size. STGCN++ is chosen as our primary skeleton-based feature extractor due to its efficiency and strong baseline performance.

\paragraph{BlockGCN~\cite{zhou2024blockgcn}} addresses two key limitations of standard GCNs in skeleton-based action recognition: (1) the decay of bone connectivity information when adjacency matrices are jointly optimized with network weights, and (2) the inefficiency in multi-relational modeling using ensemble or attention-based convolutions. To mitigate topology forgetting, BlockGCN incorporates a static topological encoding based on graph distances (e.g. shortest path distances between joint pairs) and a dynamic topological encoding via persistent homology analysis to capture action-specific skeletal dynamics. Additionally, it introduces BlockGCN, a refined graph convolution module that partitions feature channels into groups and applies spatial aggregation and projection within each group via block‑diagonal weight matrices.

\paragraph{ProtoGCN~\cite{liu2025revealing}} introduces a novel GCN-based approach for skeleton-based action recognition that explicitly enhances the model’s capacity to distinguish actions with subtly different joint dynamics. At its core is the Prototype Reconstruction Network (PRN), which learns a set of motion prototypes encoding prototypical joint-relationship patterns. Input skeleton representations are reconstructed as a weighted combination of these prototypes, thereby emphasizing fine-grained motion cues relevant for differentiating similar actions. ProtoGCN further incorporates a Motion Topology Enhancement (MTE) module that refines the graph representation via self-attention across joints and pairwise comparisons, enhancing feature richness. A class-specific contrastive learning objective encourages separation between prototype responses across action classes, reinforcing discriminative representation learning.

% \paragraph{Pose3D~\cite{duan2022revisiting}} applies 3D convolutions directly to skeleton sequences, treating pose data as an image-like representation in the temporal domain. Unlike graph-based methods, Pose3D leverages CNN-based spatiotemporal learning, benefiting from standard convolutional backbones. We use the SlowOnly variant, which maintains a single-stream architecture for temporal feature extraction, prioritizing accuracy over computational complexity.

\end{document}